\documentclass{article}
\usepackage{iclr2027_conference,times}

\usepackage{microtype}
\usepackage{graphicx}
\usepackage{subcaption}
\usepackage{booktabs}
\usepackage{hyperref}
\usepackage{url}

\definecolor{darkblue}{rgb}{0, 0, 0.5}
\hypersetup{colorlinks=true, citecolor=darkblue, linkcolor=darkblue, urlcolor=darkblue}

\usepackage{amsmath,amsfonts,bm}

\def\eqref#1{equation~\ref{#1}}

\def\1{\bm{1}}

\DeclareMathAlphabet{\mathsfit}{\encodingdefault}{\sfdefault}{m}{sl}
\SetMathAlphabet{\mathsfit}{bold}{\encodingdefault}{\sfdefault}{bx}{n}

\usepackage[utf8]{inputenc}
\usepackage[T1]{fontenc}
\usepackage{amsfonts, amsmath, amsthm}
\usepackage{amssymb} %
\usepackage{nicefrac}
\usepackage[dvipsnames]{xcolor}
\usepackage{multirow}
\usepackage{wrapfig}

\usepackage{ifsym}
\usepackage{tikz}
\usetikzlibrary{arrows.meta,calc,backgrounds,shapes.geometric,fit}
\usepackage{listings}
\usepackage{fontawesome5}
\usepackage{pgfplots}
\pgfplotsset{compat=1.18}
\usepackage{paralist}
\usepackage{placeins} %
\let\plcompactitem\compactitem \let\endplcompactitem\endcompactitem
\renewenvironment{compactitem}{\setlength{\leftmargini}{1em}\plcompactitem}{\endplcompactitem}

\newcommand{\todo}[1]{\ignorespaces}

\newcommand{\worse}[1]{\textcolor{BrickRed}{#1}}
\newcommand{\better}[1]{\textcolor{ForestGreen}{#1}}
\newcommand{\datasetname}{\textcolor{SkyBlue}{WorkArena-CF}}
\newcommand{\eg}{\textit{e.g.}}

\newcommand{\methodfullname}{\textbf{A}gent \textbf{B}ehavior as \textbf{C}ode \textbf{Agent}}
\newcommand{\methodname}{\textsc{ABCAgent}}

\title{Agent Behavior as Code: Efficient and Robust LLM Agents with Programmatic Specifications}

\author{Peng Qi, Chunliang Lyu, Gang Li$^\dagger$, Fabian Chan, Cheng Chang, Ignacio Cases$^\dagger$, Will Lu \\
Uniphore}

\iclrfinalcopy

\begin{document}

\maketitle
\lhead{}
{\renewcommand{\thefootnote}{\fnsymbol{footnote}}%
\footnotetext[2]{Work done while employed at Orby AI / Uniphore.}}

\begin{abstract}
AI agents based on foundation models (FMs) have demonstrated strong capabilities to perform complex open-ended tasks.
However, they face some common challenges in practice:
(a) agent behavior can deviate drastically even for semantically similar tasks, leading to catastrophically propagated errors;
(b) high cost and latency due to FM calls, repeated in full whenever a task recurs with different inputs;
(c) FMs' limited context and instruction following capability confine how well agents manage the ever-growing execution context and follow complex plans.
We introduce \methodfullname{} (\methodname{}), which uses a symbolic program (\eg, Python code with potential neural functions) to fully specify the agent's behavior at runtime, with a powerful FM agent editing that program for flexibility.
Behavior is thus specified without premature variable binding, and its execution is deterministic.
We evaluate \methodname{} on six agent benchmarks, two of which we construct to test how well a derived program generalizes to variants of the task it was written for.
\methodname{} matches a model-matched neural agent on GAIA and augmented GAIA, and surpasses it where robustness and long control flows matter: 98.3\% against 97.3\% on GSM-Symbolic ($p = 0.001$), 71.9\% against 47.4\% $\mathrm{Pass}^4$ on the telecom domain of $\tau^2$-bench ($p = 0.0001$), and more records written correctly at every loop length on our control-flow-augmented WorkArena benchmark.
For more parametric task families, \methodname{} is also significantly superior in efficiency.
Without authoring a new program, \methodname{} solves 92.6\% of GSM-Symbolic instances and 20.1\% of augmented GAIA variants, which yields $5.2\times$ lower latency and $7.0\times$ lower cost on GSM-Symbolic, 19\% lower cost on augmented GAIA, and $9.5\times$ lower agent latency on $\tau^2$-telecom.
\end{abstract}

\begin{figure*}[!h]
\centering
\input{figures/polished/teaser}
\caption{How \methodname{} improves over traditional FM agents. \textbf{Left:} Traditional FM agents (e.g.,~ReAct~\citep{yao2023react}) re-invoke the FM at every action step, making reruns costly when task inputs change.
\textbf{Center:} \methodname{} separates \emph{design} (task $\to$ program) from \emph{execution} (program $\to$ result).
  The program can contain neural functions (e.g., \texttt{papers()} in this example) for flexibility.
  On any new input, only the program is re-run, eliminating redundant FM calls, especially orchestration calls.
  \textbf{Right:} Error reduction against cost efficiency on six benchmarks relative to the neural agent.}
\label{fig:example}
\end{figure*}

\section{Introduction}

Foundation models (FMs) have provided a unified and generalizable interface to numerous artificial intelligence (AI) tasks spanning different modalities, giving rise to numerous agentic systems including GUI agents \citep{zheng2023seeact}, deep research \citep{huang2025deepresearchagentssystematic,zheng2025deepresearcherscalingdeepresearch}, coding agents \citep{zhang-etal-2024-codeagent, yang2024sweagent, wang2025openhands}, and interactive agents that help users complete tasks with external tools \citep{schick2023toolformer}, to name a few.

Although FM agents have demonstrated wide applicability in sophisticated tasks that are \emph{one-off}, they struggle with functioning efficiently and reliably for tasks that are \emph{recurring yet varied}.
For instance, Google's AI feature had struggled to handle ``was 1995 30 years ago''\citep{google-ai-1995-30-years-ago}, and continued to struggle on similar queries like ``was 1996 30 years ago'' after fixes.
This is challenging for purely FM-based solutions since they are shown to struggle with compositional generalization \citep{xu2024do,lake2018scan}.
Such erratic behavior would be unacceptable in real-world applications.
Ideally, the agent's behavior should be similar to a function $f(X, Y)$ for all queries like ``was $Y$ $X$ years ago'', regardless of how sophisticated the underlying task might be.
With such abstractions, it is also possible to convert some of the agent's behavior specifications into programmatic code, improving both the cost/latency by reducing unnecessary FM calls and tokens, and the transparency and controllability of the agent's behavior.

Recent work has proposed to induce programmatic skills to address these issues \citep{wang2023voyager,awm2024wang,wang2025inducing,zheng2025skillweaver, yu2026one}.
With skill induction, FM agents can write programmatic subroutines into a library that they use for subsequent tasks.
While the agent is still orchestrated by the FM, these subroutines provide determinism and efficiency for small parts of agent operations when reused.
Concurrent work~\citep{qi2026llmascode} argues that a program rather than the FM should govern an agent's control flow; our focus is instead on reusing such programs across the tasks of a family, which requires them to be re-bindable and checked for generality.

To address these challenges, various neuro-symbolic frameworks have been proposed to combine the strengths of these systems. 
However, most of these are either calling relatively simple neural functions from a purely programmatic foundation, or simply leveraging symbolic programs as tools in a larger neural agent that is still orchestrated by a foundation model (we refer the reader to \citet{saker2021neurosymbolic} for a taxonomical review of various neuro-symbolic approaches in artificial intelligence).
Neither paradigm fundamentally address the challenge of balancing the flexibility of neural agents in solving diverse tasks and the robustness and efficiency symbolic engines afford, since nesting neural and symbolic executions in a layered system is largely impossible with these formalisms.

In this work, we present \methodname{}, which tackles these challenges by organically combining FM capabilities with code execution to combine the advantages of both neural and symbolic systems.
Instead of relying on purely neural FMs to keep track of and execute complex agentic workflows, we introduce the concept of \emph{agent behavior as code (ABC)}, where the behavior of the agent is fully specified as programming code with potential calls to neural models.
This allows \methodname{} to enjoy the efficiency and robustness of programming code in orchestrating complex control flows when FM capabilities are not needed, while enjoying the flexibility of adapting to new use cases when execution failures occur, or when the neural agent determines that the requirement for the agent has changed (See Figure \ref{fig:example}).
As we will show in subsequent sections, this also allows the agent to seamlessly generalize to similar tasks with minimal to no changes to the code that defines agent logic, leveraging code to store behavioral knowledge of the AI agent to allow for efficient adaptation.

\section{Agent Behavior as Code}

In this section, we introduce the framework of agent behavior as code, with the crucial concept of separating agent design from agent execution, and detail how we implement the \methodname{} based on this framework.

\subsection{Separating Agent Design from Execution}

Before introducing the \emph{agent behavior as code} framework, we make the distinction between two phases of activities when working with AI agents, \emph{agent design} and \emph{agent execution}. 
Specifically, by \emph{agent design} we mean all activities aiming at altering the behavior of the agent to achieve the desired outcome, including but not limited to explicitly steering agent behavior through chat interactions, providing exemplars for agent tasks, providing feedback on intermediate agent outputs, etc.
On the other hand, we use \emph{agent execution} to refer to autonomous operations that the agent performs towards completing the task the user has in mind, including correct and incorrect attempts at solving the problem at design time, and subsequent application of the agent to new inputs where immediate user feedback may or may not be available.
By this definition, the design and  execution of many AI agents today are tightly coupled, where they share the same context, where agent execution logic and the execution process is coupled with the specific user inputs and examples provided at design time (with the exception of explicitly induced skills).
This coupling allows the user to intuitively design agent behavior with a responsive and flexible interface, but poses challenges when later the user needs to reuse the agent on different inputs, which we outline in subsequent paragraphs.

\paragraph{Challenge 1: Variable binding.} As agent design typically involves real or hypothetical examples of agent task input, which the agent works through with user supervision to achieve the desired behavior. 
However, when agent behavior is defined and the agent needs to generalize to new inputs, we typically rely on FMs to correctly remove the influence of example inputs provided at design time, where hallucinations can contribute to example leakage.
Classical AI systems faced the same problem: explanation-based generalization replaces the constants of a worked example with variables~\citep{mitchell1986ebg}, macro-operators and chunking store generalized solutions for later reuse~\citep{fikes1972generalizedplans,korf1985macrooperators,laird1986chunking}, and case-based reasoning adapts a retrieved solution to a new problem~\citep{kolodner1993cbr,veloso1993derivationalanalogy}.

\paragraph{Challenge 2: Robust edit of agent behavior.} Since all agent behavior is typically defined and edited in the form of agent interaction with users and the environment, this also presents a challenge when users want to edit the agent behavior at design time (\eg, by introducing if-this-then-that rules to handle a class of edge cases).
The FMs' capability to correctly follow prompt instructions in a complex context is the only lever available, and there is little verifiable guarantee that the agent design reaches the goals the user sets out to achieve or well-generalizes to unseen cases.

\paragraph{Challenge 3: Context management and execution efficiency.}
If all the design-time context is fully kept during execution time, this can introduce significant context management overhead and reduce agent execution efficiency at runtime.
Even with context compaction between design time and execution time, the aforementioned challenges still stand.

To address these challenges, we propose agent behavior as code, a framework that cleanly separates agent design from execution by leveraging programming code as the source of truth to fully specify agent behavior at execution time.
As a result, the goal of agent design is to arrive at an agentic program that captures the desired behavior with explicit generalization considerations and no premature variable binding, focusing on specifying agent behavior in general rather than what is the expected result in a specific case.
This addresses all aforementioned challenges by 1) explicitly define and use unbound input variables in the agentic program; 2) use code logic to reflect agentic behavior to ensure edit lineage and avoid regression; 3) delegate expensive LLM-based orchestration to much more efficient code executors when possible, which is not only efficient and robust, but also enjoys unbounded execution context thanks to modularization, variable scoping, and nesting natively supported by programming languages (\eg, a for loop that runs $n$ times would not require $O(n)$ context / memory, an observation concurrent work also makes~\citep{qi2026llmascode}).

\subsection{Implementation of \methodname{}}\label{sec:impl}

\begin{figure}[t]
\centering
\resizebox{0.8\linewidth}{!}{\input{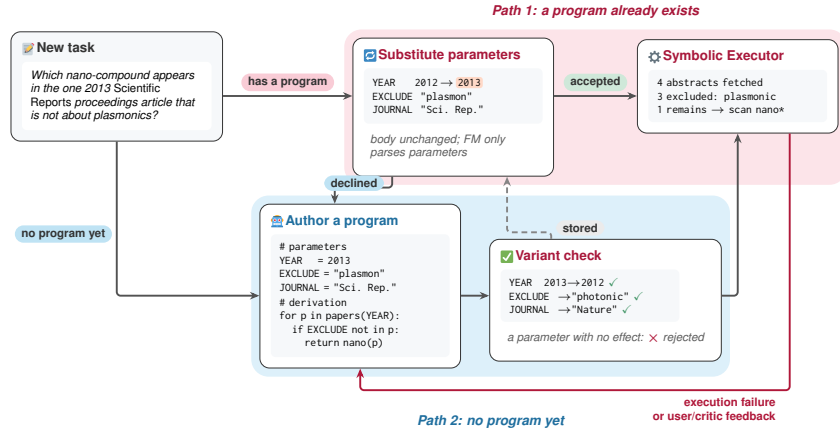}}
\caption{\methodname{} processes new tasks by authoring programs from scratch, or reusing them from previously solved similar tasks.
In the latter case, \methodname{} attempts to substitute parameter values into the existing program (with a lightweight FM call for parsing only when necessary), before resorting to editing the program if substitution fails.}
\label{fig:loop-diagram}
\end{figure}

To generalize to unseen inputs, an \methodname{} program declares the parameters it can be re-bound on up front, separating the values a particular task supplies from the logic that consumes them.
A program is therefore applied to a later task by supplying values rather than by regenerating its logic.

The authoring loop itself is a standard code-generation cycle.
The FM Designer analyzes the task with the current program, execution state and feedback from previous executions (if any) to propose a program that addresses the task requirements.
It then hands the proposed program to the Symbolic Executor for end-to-end validation.
When execution fails or when an LLM-based critic detects an issue with the program, the outcome is returned to the FM Designer for revision.
During authoring, we ask the FM Designer to propose variants of the task at hand, and solve them alongside the original task with its program.
These variants are then used to check the generality of the program through execution before it is accepted; if the program fails to change its behavior appropriately for the variants, it is rejected.
This check is executable and cheap, and it rejects programs whose generality is merely declared but not implemented.
This check is akin to metamorphic testing~\citep{chen1998metamorphic,segura2016metamorphicsurvey}: each variant asserts how the output should change when a parameter does, so a program that ignores the parameter fails it.
It is related to validating generated code against generated tests~\citep{chen2022codet} and to property-based testing~\citep{claessen2000quickcheck}; see \citet{zheng2026metamorphicllmsurvey} for a survey of metamorphic testing with language models.

When \methodname{} encounters a new task, it begins by authoring a fresh program from scratch.
Once a program is drafted and a subsequent task in the same problem family is encountered, \methodname{} will first attempt to apply the existing program to the new task by substituting the new task's parameter values.
Should this attempt fail, the FM Designer is invoked to revise the program so that it can handle the new task without regressing on previously solved tasks.
This allows the system to leverage existing knowledge efficiently, only invoking the FM Designer when necessary to accommodate new variations.

For illustration, consider the parameterized literature-search task of Figure~\ref{fig:loop-diagram}.
A purely FM-based agent would repeat the entire search process with FM-orchestrating calls due to the nested, iterative nature of this task; in contrast, \methodname{} can reuse generally applicable programs to handle new tasks more efficiently.

\subsection{Implications for State and Memory Management}

\methodname{}'s neural-symbolic architecture fundamentally changes agent memory management by structuring it around the principles of program execution.

\begin{compactitem}
    \item \textbf{Short-Term Knowledge (Variable Management):} Instead of injecting all knowledge back into a flat text context, agent state and short-term memory are managed as program variables. When the orchestrator executes \texttt{x = 5}, the variable \texttt{x} is stored with its value and type. This state is precise, typed, and requires no interpretation. It avoids the ambiguity and context-length limitations of trying to remind an FM that ``we previously established a value of 5 for the concept x.''

    \item \textbf{Execution Progress (Execution Frames and Scoping):} The agent's progress is not merely a transcript of past actions but a structured call stack. When the orchestrator enters a for loop or a function, it creates a new execution frame with its own local scope. This allows the agent to handle complex, nested tasks without losing track of its position in the overall plan. For instance, the orchestrator, not the FM, is responsible for tracking that it is on iteration 2 of 3 in the loop from our example. This robustly manages execution state, preventing the kind of drift or forgetfulness common in long-running, purely neural agents. Furthermore, variable scoping is handled naturally by the symbolic engine, preventing state pollution between different subtasks or loop iterations. This structure inherently supports modularity, nesting, and even recursion, enabling a level of complexity that is intractable with a flat memory model.
\end{compactitem}

\section{Experiments}\label{sec:experiments}

In this section, we demonstrate the efficacy and efficiency of \methodname{} on a number of diverse agent benchmarks.

\subsection{Experimental Setup}

\paragraph{Benchmarks \& Metrics.} We evaluate \methodname{} on the following agent benchmarks: 
1) \textbf{GAIA} \citep{mialon2023gaiabenchmarkgeneralai} is a collection of general AI agent tasks that require multistep autonomous reasoning and tool use for textual question answering.
The dataset is divided into L1, L2, and L3 subsections for increasing difficulty, where more tools and reasoning steps are required for higher levels.
We use the development set for evaluation.
2) \textbf{WorkArena} \citep{workarena2024, boisvert2024workarenacompositionalplanningreasoningbased} features realistic, enterprise-focused web tasks on the ServiceNow platform.
WorkArena is similarly divided by difficulty levels; we evaluate on the L1 subset.
3) \textbf{GSM-Symbolic} \citep{mirzadeh2024gsmsymbolic} augments 100 grade-school math word problems with symbolic templates, and samples 50 instances from each template with varying names and numerical values while keeping the underlying reasoning fixed.
4) \textbf{$\tau^2$-bench} \citep{barres2025tau2bench} evaluates conversational agents in customer-service domains, where the agent must follow a domain policy while helping a simulated user.
We use the telecom and airline domains, and report the benchmark's $\mathrm{Pass}^k$ metric, which requires all $k$ trials to succeed ($k\in\{1,4\}$).
For all of these benchmarks, we adopt the official evaluation metric that calculates average task success rate on each benchmark or subset thereof.

To stress test \methodname{} on generalization and efficacy of handling complex tasks, we further augment existing benchmarks.\footnote{We will release the implementation of \methodname{} and both augmented benchmarks upon acceptance.}
For GAIA, we generate questions that slightly modify the original ones in the development set, and annotate corresponding answers to these modified questions with Co-Sight \citep{zhang2025cosight}, a high-accuracy system on the GAIA leaderboard with some human verification.
We then test how well the agent behavior program that \methodname{} derives from the original question generalizes to these new tasks.
For WorkArena, we compose the original L1 tasks into more complex ones with control flows (e.g., if/else branching or for loops) that the agent must follow to successfully complete these tasks.
We created 18 augmented tasks in total for this \datasetname{} benchmark, 8 involving loops and 10 involving branching.
For both augmented datasets, we use evaluation metrics consistent with the original dataset for task success rate calculation.
We refer the reader to Appendices \ref{appdx:gaia_augmented} and \ref{appdx:workarena_augmented} for details about these augmented datasets.

To further illustrate the execution efficiency of \methodname{}, we also compare the time it takes a neural agent and \methodname{} to finish tasks on these benchmarks, the time it takes for execution without redesign, the cost of FM calls, the share of tasks solved without authoring a program, and other diagnostic metrics, if applicable.
We refer the reader to Appendix~\ref{appdx:protocols} for further details about the evaluation protocols, including cost accounting and statistical tests.

\paragraph{Baselines.} We compare \methodname{} to state-of-the-art approaches on each benchmark:
for public reference, we compare against the leading agents on the same data split that we evaluate on for each benchmark.
We also implement our own version of a purely neural agent for comparison, which leverages the same Claude Sonnet 4.6 model as its backbone and Claude Code as its agent design -- but unlike \methodname{}, it does not store or reuse agent behavior programs.
This neural baseline is more important especially for GAIA tasks, where targeted harness engineering can make a significant difference in performance.

\subsection{Main Results}

\paragraph{\methodname{} is as accurate as neural agents, and more robust in complex and symbolic-like tasks.}
As can be seen in Table \ref{tab:main_results}, on GAIA, augmented GAIA, WorkArena, and $\tau^2$ airline domain, we note that the accuracy of \methodname{} is statistically indistinguishable from that of a model- and harness-matched neural agent.
For GAIA and Augmented GAIA,this accuracy match holds across different difficulty levels and task variations (see Appendix Table \ref{tab:gaia_results}).
This shows that we do not lose accuracy by specifying agent behavior through explicit programs compared to relying on neurally-orchestrated agents.
Furthermore, on GSM-Symbolic and $\tau^2$ telecom, two tasks that are more amenable to symbolic generalization, \methodname{} demonstrates a clear advantage over neural agents.
On $\tau^2$-bench, \methodname{} is also more stable across repeated trials: its Pass$^1$ to Pass$^4$ drops from $81.6\%$ to $71.9\%$ on telecom and from $68.5\%$ to $60.0\%$ on airline, more modest compared to the drops of $60.8\%$ to $47.4\%$ and $68.5\%$ to $58.0\%$ for the stock neural agent (Appendix Table~\ref{tab:tau2}).
On \datasetname{}, even when full task success rate is comparable, \methodname{} consistently writes more database records correctly than the neural agent at every loop length (Appendix Figure~\ref{fig:wacf_sweep}a).
This highlights \methodname{}'s robustness in scenarios where explicit programmatic reasoning is beneficial.

\paragraph{\methodname{} is more cost- and time-efficient, especially with reuse or on tasks with complex control flows.}
Table~\ref{tab:main_results} also shows that \methodname{} is at least as cost- and time-efficient as the neural agent on most benchmarks, and substantially more so where its programs are reused.
On GSM-Symbolic and $\tau^2$-bench, \methodname{} is noticeably more cost- and time-efficient than the neural agent at the same time.
We note that these savings do not come from cheaper inference for FM calls necessary to solve the task, which costs about as much for \methodname{} as for the neural agent; instead, they come from the $92.6\%$ of GSM-Symbolic tasks and $20.1\%$ of augmented GAIA tasks that \methodname{} solves without authoring, by rebinding the parameters of an existing program (see Section~\ref{sec:analysis} for detailed analysis).

\begin{table*}[t]
\centering
\small
\setlength{\tabcolsep}{5pt}
\resizebox{\textwidth}{!}{%
\begin{tabular}{@{}lccccccc@{}}
\toprule
\multirow{2}{*}{System} & \multicolumn{2}{c}{GAIA} & \multicolumn{2}{c}{WorkArena} & GSM-Symbolic & \multicolumn{2}{c}{$\tau^2$-bench ($\mathrm{Pass}^4$)} \\
\cmidrule(lr){2-3} \cmidrule(lr){4-5} \cmidrule(lr){7-8}
 & base (165) & augmented (378) & L1 (31) & \textcolor{SkyBlue}{CF} (42) & (5000) & telecom (114) & airline (50) \\
\midrule
Reference & 74.6 & --- & 42.7 & --- & 94.9 & 29.0 & 42.0 \\
Neural agent & \textit{76.4} & \textit{59.5} & \textit{84.9} & 58.3 & 97.3 & 47.4 & 58.0 \\
\methodname{} & 72.7 & 57.1 & 62.4 & \textit{64.3} & \textbf{98.3} & \textbf{71.9} & \textit{60.0} \\
$p$-value & 0.377 & 0.281 & 0.041 & 0.453 & \textbf{0.0011} & \textbf{0.0001} & 1.000 \\
\midrule
Cost efficiency & 0.90$\times$ & \better{1.24$\times$} & \worse{0.41$\times$} & 1.02$\times$ & \better{6.97$\times$} & \better{12.93$\times$} & \better{5.66$\times$} \\
Speedup & 0.85$\times$ & \better{1.33$\times$} & \worse{0.36$\times$} & 0.92$\times$ & \better{5.21$\times$} & \better{9.50$\times$} & \better{4.08$\times$} \\
\bottomrule
\end{tabular}%
}
\caption{Task success rate (\%) and efficiency of \methodname{} relative to model- and harness-matched neural agent on six agent benchmarks.
Bold marks a statistically significant difference ($p<0.01$) and italics the stronger system otherwise; cost efficiency and speedup are the neural agent's cost and time divided by those of \methodname{} on the tasks where both systems produced a result, highlighted when the deviation is $\ge 20\%$.
The reference row is the strongest published result we are aware of, with --- where none exists (see Appendix~\ref{appdx:protocols} for details).}
\label{tab:main_results}
\end{table*}

\section{Analysis}\label{sec:analysis}

In this section, we perform detailed analysis of the mechanisms and behaviors of \methodname{}.

\paragraph{Program reuse and variable rebinding drive the efficiency of \methodname{}, and the variant check makes its programs generalizable.}
As can be seen in Table~\ref{tab:mechanism}, while variant check introduces moderate overhead in cost, it helps improve the generalizability and correctness of the programs authored by \methodname{} especially on \datasetname{}.
Meanwhile, sharing programs across similar tasks results in the biggest efficiency gains for \methodname{} by reducing the need to author new programs for each task.
We also note that these components are not additive: on GSM-Symbolic, family reuse without cheap rebinding is more expensive than no reuse at all (Appendix Table~\ref{tab:ablation_gsm}), since every instance still goes through the authoring loop, only now with the family's program in its context.

\begin{table*}[t]
\centering
\small
\setlength{\tabcolsep}{3pt}
\newcommand{\cm}{\checkmark}
\begin{tabular}{@{}l@{\hskip 6pt}ccc@{\hskip 6pt}ccc@{\hskip 6pt}ccc@{}}
\toprule
 & \multicolumn{3}{c}{Augmented GAIA L1 (129)} & \multicolumn{3}{c}{GSM-Symbolic (1000)} & \multicolumn{3}{c}{\datasetname{} (18)} \\
\cmidrule(r{6pt}){2-4} \cmidrule(r{6pt}){5-7} \cmidrule{8-10}
Setting & SR & Free & Cost eff. & SR & Free & Cost eff. & SR & Records & Cost eff. \\
\midrule
Neural agent & 62.0 & 0.0\% & 1.00$\times$ & 99.7 & 0.0\% & 1.00$\times$ & 76.5 & 53.3 & 1.00$\times$ \\
\midrule
\methodname{} & 58.9 & 21.7\% & \better{1.27$\times$} & 99.1 & 96.0\% & \better{11.91$\times$} & \textit{83.3}$^{\S}$ & \textit{63.9} & 1.04$\times$ \\
\quad $-$ reuse \& rebind & \textit{62.8} & 0.0\% & 1.02$\times$ & 99.4 & 0.0\% & 0.96$\times$ & 77.8 & 63.3 & 0.80$\times$ \\
\qquad $-$ variant check & 56.6 & 0.0\% & \better{1.30$\times$} & \textit{99.8} & 0.0\% & 0.97$\times$ & 68.8 & 0.0\rlap{$^{\ddagger}$} & \better{1.26$\times$} \\
\bottomrule
\end{tabular}
\caption{Ablation of \methodname{}'s components.
We use the L1 subset of Augmented GAIA and 20\% of GSM-Symbolic for this experiment.
\emph{Free} is the share of tasks solved without authoring a program, and \emph{Records} is the percentage of records written correctly on the loop tasks ($N=30$).
$^{\S}$On \datasetname{}, cheap rebinding never took effect, so this setting behaves as family reuse with the variant check alone.
$^{\ddagger}$Without the variant check, the replayed programs commit records with wrong field values, including one program that ignores its per-record input entirely (details in Appendix~\ref{appdx:ablations}).}
\label{tab:mechanism}
\label{tab:amortization}
\label{tab:binding}
\end{table*}

\begin{figure*}[t]
\centering
\input{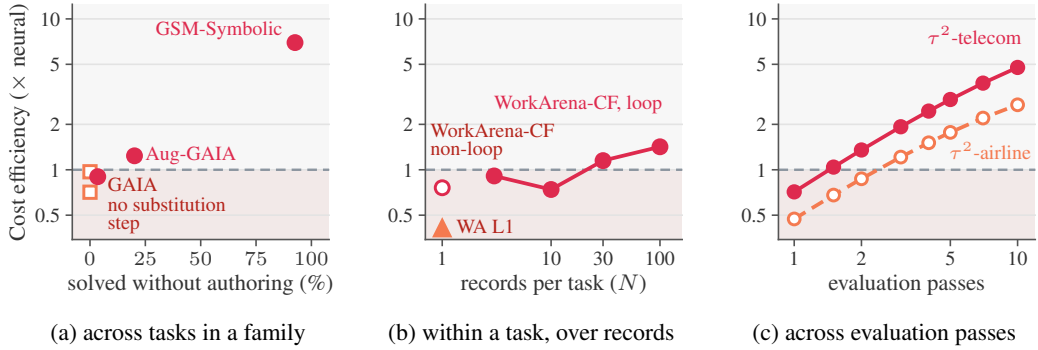}
\caption{Cost efficiency of \methodname{} relative to the neural agent under three forms of amortization: (a) reusing a program across the tasks of a family, (b) reusing a loop body, and (c) evaluating a policy repeatedly.
The open squares in (a) are \methodname{} on GSM-Symbolic without rebinding.}
\label{fig:free_rate}
\end{figure*}

We find that these savings take three different forms (Figure~\ref{fig:free_rate}): across the tasks of a family, cost efficiency grows with the share of tasks that \methodname{} solves without authoring; within a single task, a program-driven loop amortizes the authoring cost over the records it processes; and on $\tau^2$-bench, one authored policy is evaluated repeatedly and pays back its authoring cost in fewer than three passes.
This also explains the below-parity cost efficiency and speedup on WorkArena L1 in Table~\ref{tab:main_results}.
WorkArena L1 affords none of these, since each of its tasks is a one-off that is neither repeated nor looped over, so \methodname{} pays the authoring cost in full without ever amortizing it.
Its lower nominal accuracy there has a different cause: most of \methodname{}'s failures on WorkArena L1 end with the browser on a page other than the one the benchmark's validator reads, such as a catalog order that never reaches its confirmation page (Appendix~\ref{appdx:analysis}), which we suspect is due to the authored program not checking the page state it leaves behind.

\begin{table}[t]
\centering
\small
\setlength{\tabcolsep}{5pt}
\begin{tabular}{@{}lrccc@{}}
\toprule
Transfer setting & pairs & Reuse & Control & $p$ \\
\midrule
\multicolumn{5}{@{}l}{\emph{GSM-Symbolic}} \\
\quad Forward ($j>i$) & 7\,640 & \textbf{0.958} & 0.842 & $\mathbf{<10^{-4}}$ \\
\quad Backward ($j<i$) & 7\,360 & \textbf{0.963} & 0.826 & $\mathbf{<10^{-4}}$ \\
\midrule
\multicolumn{5}{@{}l}{\emph{Augmented GAIA}} \\
\quad Forward ($j>i$) & 235 & \textit{0.332} & 0.302 & 0.410 \\
\quad Backward ($j<i$) & 239 & \textbf{0.469} & 0.305 & \textbf{3.5e-5} \\
\quad Backward, target known-passing & 145 & \textbf{0.703} & 0.462 & \textbf{5.1e-5} \\
\bottomrule
\end{tabular}
\caption{Fraction of task pairs within a family on which a program stored for one task (indexed by $i$) solves the other (indexed by $j$), for unseen (forward) and earlier (backward) tasks, with and without program reuse.
On GSM-Symbolic, parameters are rebound with one FM call (Appendix~\ref{appdx:protocols}); bold marks $p<0.01$ and italics the stronger result otherwise.}
\label{tab:transfer}
\end{table}

\paragraph{Program reuse retains strong backward compatibility within a family while moderately improving forward generalization.}
We also find that program reuse substantially improves backward compatibility within a family of tasks on both GSM-Symbolic and Augmented GAIA, retaining the ability to solve earlier tasks even as new ones are added, allowing for more reliable cumulative learning via robustly evolving the agent's behavior (see Table~\ref{tab:transfer}).
This is because a stored program is revised under the requirement that it still solves the tasks it has solved before.
In contrast, neural agents and programs authored independently for each task carry no such guarantee.
On the forward direction, we find that reuse also improves the program's ability to solve yet unseen tasks within the same family, substantially on GSM-Symbolic but only nominally on Augmented GAIA, where the effect is not as pronounced as in the backward direction, because the forward tasks may introduce new requirements that were not present in the previously solved tasks.

On GSM-Symbolic, where variations among the instances of a family are typically in names and numbers over one fixed derivation, a stored program transfers to nearly all of its family, whereas programs authored one task at a time fail noticeably more often on their siblings.
This makes GSM-Symbolic more amenable to few-shot learning for both task logic and shared interface, as also evidenced by the high ``free'' program rate in Table~\ref{tab:mechanism}.

Hence, what reuse primarily provides is a stronger guarantee against regression, with improved forward generalization when new variants conform better to the patterns of seen tasks, although little exposure is needed before a stored program covers most of its family. 
We refer the reader to Appendix~\ref{appdx:transfer} for further analysis.

\paragraph{The generalization of \methodname{} does not come from observing past runs alone.}
A natural question is, does \methodname{} generalize simply because it observes the transcripts of past runs? 
To answer this, we give a neural agent on $\tau^2$-telecom the same transcripts as context (Table~\ref{tab:tau2_ladder}).
While the neural agents benefit modestly from this information about prior tasks, \methodname{} benefits considerably more, which suggests that it is specifying behavior as a program, rather than as prose context, that makes past runs useful.
We additionally report an oracle ceiling that reveals the reward of the task being authored, to which our main configuration has no access, to indicate how much further \methodname{} could benefit from richer adaptation signals.

\paragraph{Other analyses.}
We further find that programs can helpfully defer to \methodname{} nested-calls only where it resists code, that errors on \datasetname{} do not accumulate with loop length, and that when rebinding fails, the reason is legible in the authored program; see Appendix~\ref{appdx:analysis} for details.

\begin{table}[t]
\centering
\small
\begin{tabular}{@{}llcc@{}}
\toprule
System & Behavior specification & Adaptation signal & $\mathrm{Pass}^1$ \\
\midrule
Neural agent & prompt only & none & 0.603 \\
Neural agent $+$ traces & prompt $+$ prior transcripts & task transcripts & 0.671 \\
\methodname{} & symbolic program & task transcripts & 0.789 \\
\quad $+$ reward at authoring & symbolic program & transcripts $+$ reward & 0.943 \\
\midrule
\multicolumn{3}{@{}l}{\emph{Control}: neural agent, transcript injection disabled} & 0.638 \\
\bottomrule
\end{tabular}
\caption{Prompt-based neural agents and \methodname{} adapting in a continual learning setting on $\tau^2$-telecom. Revealing the reward at authoring is an oracle ceiling.}
\label{tab:tau2_ladder}
\end{table}

\section{Background and Related Work}

\paragraph{Neuro-symbolic AI.}
The effective combination of neural and symbolic systems has a rich history~\citep{garcez2023neurosymbolic3wave,saker2021neurosymbolic,hitzler2022neurosymbolicapproaches,bhuyan2024neurosymbolicsurvey,belle2024neurosymbolicagents}.
\citet{kautz2022third} highlights prominent paradigms, among which \emph{Symbolic[Neuro]} denotes a symbolic program that calls neural functions (\eg, AlphaGo's MCTS over a neural value function), and \emph{Neuro[Symbolic]} denotes a neural system with embedded symbolic reasoning (\eg, an LLM invoking a theorem prover).
\methodname{} proposes a new paradigm $\mathit{NS} = \textsc{Neural}[\textsc{Symbolic}[\mathit{NS}]]$, a recursive definition: the outermost $\textsc{Neural}[\cdot]$ is the FM Designer that generates and edits programs, while $\textsc{Symbolic}[\mathit{NS}]$ is the executor---a deterministic engine that may invoke neural tools or recursively call the FM Designer itself to handle sub-problems, yielding the nested $\mathit{NS}$.
This enjoys the plasticity of neural systems alongside the efficiency, robustness, and out-of-distribution generalization of symbolic execution.
Representing agent behavior as code also improves transparency, interpretability, and controllability; we refer the reader to prior work with detailed discussions on this topic~\citep{hamilton2022neurosymbolicnlp,tocchetti2025robustness}.

\paragraph{Neuro-symbolic methods for specific tasks.}
Neuro-symbolic systems have been applied across many tasks, illustrating Kautz's taxonomy of collaboration schematics.
In \emph{Symbolic[Neuro]}, \citet{silver2016alphago} and \citet{silver2018alphazero} combine Monte Carlo Tree Search (MCTS) with deep neural value and policy functions; \citet{manhaeve2018deepproblog} embed neural networks as callable predicates within probabilistic logic programs; and \citet{mitchener2022detect} learn an Answer Set Programming (ASP) meta-policy via Inductive Logic Programming (ILP) over deep reinforcement learning (RL) sub-policies.
In \emph{Neuro$\|$Symbolic}, \citet{mao2019nscl} parse visual scenes into symbolic representations via neural perception, \citet{chaudhury2021neurosymbolicpolicy} learn symbolic action policies via differentiable rule learning from neural text encodings,and \citet{liang2016neuralsymbolicmachines} learn to generate executable programs against a knowledge base from weak supervision.
In \emph{Neuro\_\{Symbolic\}}, \citet{andreas2015neuralmodulenetworks} assemble a composition of neural modules from a parsed layout.
In \emph{Neuro[Symbolic]}, \citet{zheng2025jarvis} embed a symbolic planner inside a large language model (LLM)-driven embodied agent, \citet{li2024neurosymbolicmath} combine LLMs with symbolic solvers for math data generation, and---as discussed below---agents that execute generated code as actions also fall in this paradigm.
Graph-based approaches further combine graph neural networks (GNNs) with symbolic relational structures for knowledge-grounded reasoning~\citep{hitzler2022neurosymbolicapproaches,saker2021neurosymbolic}.
In contrast, the FM Designer in \methodname{} can implement \emph{any} symbolic structure that best fits the task, and its use of functionally versatile programmatic code paired with a general-purpose coding agent makes it applicable to arbitrary agentic tasks.

\paragraph{Agents that generate and execute code.}
Early neural program execution~\citep{reed2016npi} predates the LLM era; LLMs trained on code~\citep{chen2021codex} then enabled applications such as language-model programs for embodied control~\citep{liang2023codeaspolicies}, ambiguous-to-executable program synthesis~\citep{zelikman2022parsel} and program-aided reasoning, where the model writes a program and an interpreter executes it~\citep{gao2022pal,chen2022programofthoughts}.
\citet{yang2024codewand} survey how code empowers LLMs as agents; \citet{xu2024lemur} show that jointly training on natural language and code yields stronger language agent backbones.
CodeAct~\citep{wang2024executable} and REPL-Plan~\citep{liu2025replplan} direct agents to emit executable code snippets as actions, while skill-induction approaches~\citep{wang2023voyager,awm2024wang,wang2025inducing,zheng2025skillweaver,yu2026one} extract reusable programmatic subroutines from execution traces.
Among these, CodeAct and skill-induction agents are the most relevant to \methodname{}; like other \emph{Neuro[Symbolic]} systems, all place the FM as the outer neural controller that invokes symbolic code.
However, both employ code only in the \emph{periphery}: CodeAct generates transient snippets discarded after each task, and skill-induction agents still rely on the FM to orchestrate top-level control flow, with neither lifting variable scoping to the full agent level.
\methodname{} differs by making the symbolic program the \emph{central artifact} that fully specifies agent behavior under the recursive $\mathit{NS} = \textsc{Neural}[\textsc{Symbolic}[\mathit{NS}]]$ paradigm---elevating code from a tool called by the FM to a persistent specification managed by it.
Another line of work has the FM write the agent itself: Meta Agent Search programs new agent designs into an archive~\citep{hu2024adas}, AFlow searches over workflows represented as code~\citep{zhang2024aflow}, Trace optimizes agent code from execution feedback~\citep{cheng2024trace}, and DSPy compiles declarative LM pipelines into self-improving programs~\citep{khattab2023dspy}.
These methods search offline for a general agent design scored on a benchmark, whereas \methodname{} authors a program for one task family and reuses it by rebinding its parameters.
Lifting the constants of a worked example into reusable parameters is also the goal of abstraction learning over program libraries~\citep{ellis2020dreamcoder,grand2023lilo,stengeleskin2024regal}.
In concurrent work, \citet{qi2026llmascode} likewise let a program rather than the FM govern an agent's control flow, and \citet{pedada2026sharedpersistentmemory} reuse generated programs across sessions by refreshing their data without a model call.
Our focus differs in that every \methodname{} program declares re-bindable parameters and passes an executable check of its generality before it is stored, and in that we measure how the resulting savings amortize over task families on six benchmarks.

\section{Conclusion}
In this paper, we presented \methodname{}, which specifies the behavior of an FM agent as a symbolic program that the FM designs and edits, thereby separating the design of agent behavior from its execution.
A program \methodname{} authored on one task can be reused across other tasks of the same family, so that when tasks recur with varied inputs, \methodname{} is as accurate as a neural agent or more, at a fraction of its cost and latency.
We believe that treating agent behavior as code opens several directions for future work, such as agent behavior engineering and generalization limitations of code-driven agents.

\subsection*{AI use statement}
In this work, we used generative AI tools for generating synthetic data sets, cleaning and reformatting datasets, implementing methods, designing and providing feedback on experiments, interpreting results, and refining hypotheses and the conceptual framework, including the decomposition of amortization into three mechanisms shown in Figure~\ref{fig:free_rate}.
The main hypotheses and ideas of this work are the authors' own.
The variant tasks of augmented GAIA were largely generated with generative AI tools, which also helped correct annotation errors in the dataset; these corrections were validated by the authors.
The tasks of \datasetname{} were largely proposed by the authors, and generative AI tools were used to develop the code that implements them.
We have not used generative AI tools to formulate mathematical claims, to provide critical ingredients for proving mathematical claims, or to assist in the writing of proofs, and translation and qualitative or thematic data analysis are not applicable to this work.
Additionally, we used generative AI tools for creating and editing software code, creating and modifying scientific figures, drafting parts of the paper, identifying and summarizing relevant literature, searching for information, formatting references, and brainstorming.
We have reviewed all AI-assisted work.
AI-generated code was tested and verified by one or two authors, depending on the portion of the work.
AI-suggested references were checked against their arXiv or Crossref records and read by an author before being cited.
We take responsibility for the final content of this work, including text, claims or artifacts produced with the aid of generative AI.

\subsection*{Ethics statement}
This work does not involve human subjects, and none of our benchmarks contain personal data: GAIA, WorkArena, GSM-Symbolic and $\tau^2$-bench are public benchmarks, the users of $\tau^2$-bench are simulated, and augmented GAIA and \datasetname{} are derived from GAIA and WorkArena as described in Appendices~\ref{appdx:gaia_augmented} and~\ref{appdx:workarena_augmented}.
All web and enterprise tasks run on the benchmarks' own sandboxed environments.
Agents that act autonomously on the web or in enterprise software can cause harm when their actions are wrong, and \methodname{} replays a stored program without a model call, which can repeat a wrong action across many inputs.
We mitigate this by checking a program on task variants before it is stored, and we recommend that deployments keep such checks, together with human review of stored programs for consequential actions.

\subsection*{Reproducibility statement}
Section~\ref{sec:impl} describes \methodname{}, and Section~\ref{sec:experiments} and Appendix~\ref{appdx:protocols} describe the baselines, the model used by both systems, the evaluation protocols, how cost and time are measured, and the statistical tests.
The construction of augmented GAIA and \datasetname{} is described in Appendices~\ref{appdx:gaia_augmented} and~\ref{appdx:workarena_augmented}.
We will release the implementation of \methodname{} and both augmented benchmarks upon acceptance.

\bibliography{iclr2027_conference}
\bibliographystyle{iclr2027_conference}

\clearpage

\appendix
\section{Limitations} \label{appdx:limitations}
We acknowledge several limitations of our study, which we discuss in turn below.

\paragraph{Implementation backend and run-to-run variance.}
\methodname{} is implemented on a single agent harness, and its components only take effect on that harness, so the results reported here might not transfer directly to other agent frameworks.
Agent evaluations are also noisy from run to run: on the L1 tasks of augmented GAIA, $19.4\%$ of tasks change outcome when only the order of sibling tasks changes, and on \datasetname{} the raw success rate of an identical configuration ranges from $3$ to $5$ of $8$ composites across repeated draws.
We therefore report replicated or paired comparisons wherever possible, and treat single-run differences of a few points as uninformative.

\paragraph{Ground truth of augmented GAIA.}
The answers to the augmented GAIA tasks are produced by an LLM agent through majority voting and validated by the authors where errors were found (Appendix~\ref{appdx:gaia_augmented}), so it cannot be guaranteed that every label is correct, especially for tasks that turned out to be infeasible.
Because both systems are scored against the same labels, label errors affect absolute success rates more than the comparison between them, but they may still mask small differences.

\paragraph{Contamination.}
GAIA and the GSM8K problems underlying GSM-Symbolic were public before the training cutoff of the model we use, so the model may have seen some of these tasks or their answers.
Our comparisons are less exposed to this than absolute scores, since the neural agent and \methodname{} share the same model, and the variants in augmented GAIA and \datasetname{} are our own and have not been released; nevertheless, we cannot rule out that contamination favors one system over the other.

\paragraph{Single model family.}
All agents in our experiments use Claude Sonnet~4.6, both as the neural agent and as the FM Designer of \methodname{}, so we have not established how the results change with other foundation models, including smaller models for which authoring a correct program might be harder.

\paragraph{Dependence of the efficiency gains on reuse.}
The cost and latency savings of \methodname{} depend on how much its programs are reused, whether across the tasks of a family, over the records processed by a loop, or over repeated evaluation passes (Figure~\ref{fig:free_rate}).
Below the corresponding break-even point \methodname{} is more expensive than the neural agent: it costs about $2.4\times$ as much on the one-off tasks of WorkArena L1, is less cost-efficient on \datasetname{} at $10$ records per task, and needs about $1.4$ and $2.4$ evaluation passes to pay back its authoring cost on the two $\tau^2$-bench domains.
Our efficiency claims therefore apply to recurring tasks, and might not be generally applicable to workloads in which most tasks are seen only once.

\paragraph{Sample size.}
Several of our benchmarks are small: WorkArena L1 has $33$ tasks, \datasetname{} has $8$ loop composites of which $6$ write records, and the hardest level of GAIA has $26$ tasks.
At these sizes paired tests have little power, so a non-significant difference there should be read as underpowered rather than as evidence of no effect, and per-level results as descriptive.

Taken together, our results should not be treated as a universal statement about programmatic agents, but rather as evidence on recurring tasks under one model and one harness; a broader evaluation across models, harnesses and task distributions warrants further investigation, which we leave to future work.

\section{Experimental Details} \label{appdx:protocols}
This section describes the evaluation protocols behind the results in the main text.

\paragraph{Models and reference results.}
All systems use Claude Sonnet~4.6 and are model-matched within each benchmark; on $\tau^2$-bench, the neural baseline is the benchmark's own stock agent, which is also the agent that public $\tau^2$-bench leaderboards vary the model under.
The reference rows give the strongest published result we are aware of, which is not directly comparable to our rows: the WorkArena L1 protocol samples $10$ seeds per task under a $15$-step cap where we run three replicates per task, the GSM-Symbolic figure is obtained by direct prompting rather than by a tool-using agent, the $\tau^2$-bench figures use \texttt{o4-mini}, the strongest of the four models evaluated by \citet{barres2025tau2bench}, under the stock agent, and the GAIA figure is the strongest generalist agent on the HAL leaderboard \citep{hal2025leaderboard}, which evaluates the same 165-question validation split, whereas the public GAIA leaderboard reports on the test split.

\paragraph{Scoring.}
On augmented GAIA, both systems are compared on the same 378 child tasks; \methodname{} additionally solves the 165 parent tasks from which programs are reused, which the neural agent does not need.
\datasetname{} comprises $42$ task configurations, namely $10$ non-loop tasks and $8$ loop composites at four loop lengths ($N \in \{3, 10, 30, 100\}$); in Table~\ref{tab:main_results}, all $42$ configurations are scored by task success, with loop composites adjudicated against the live instance rows, the four repeated draws at $N\!=\!3$ averaged, and re-run tasks superseding their original run, while the fraction of records written correctly is reported per loop length.
Three \datasetname{} runs flagged by an internal mechanism check, rather than by an environment fault, are kept with reduced denominators.
On $\tau^2$-bench, each task is run for four trials; $\mathrm{Pass}^1$ is the success rate of a single trial, and $\mathrm{Pass}^4$ counts a task as solved only if all four trials succeed.
On WorkArena L1, both systems run with the same three seeds, and we compare the $31$ tasks whose instances match in every replicate; the other two draw identifiers that the seed does not control.

\paragraph{Cost and latency.}
Costs are computed from the token usage of each model call, including the short calls that bind a new task onto a stored program and verify its result; a task solved by cheap rebinding makes two such calls and no others, so we call it \emph{free} in the sense that it runs no authoring loop.
On GSM-Symbolic these calls add \$$18.01$ over $5{,}000$ tasks.
Their cost was not recorded on augmented GAIA, base GAIA and \datasetname{}; from the calls recorded on GSM-Symbolic, they would lower the cost efficiency of \methodname{} on augmented GAIA by at most about $3\%$, only widen the neural agent's lead on base GAIA, and on \datasetname{} could offset \methodname{}'s small cost advantage entirely, so we do not claim a cost difference on \datasetname{}.
Their time was not logged, so we measured it by rerunning the binding and verification of $20$ randomly sampled free GSM-Symbolic tasks, which take $3.6$\,s per task ($95\%$ CI $[3.2, 4.0]$\,s), and added it to \methodname{}'s time; this lowers its speedup on GSM-Symbolic from $8.2\times$ to $5.2\times$ ($5.0$--$5.4\times$ over the CI).
These calls cost little but take about as long as a short FM call each, which is why they reduce the speedup much more than the cost efficiency.
Because the rerun was serial whereas the original run was concurrent, $5.2\times$ is more likely an overestimate than an underestimate.
On augmented GAIA, adding the binding and verification calls, measured at $5.4$\,s per accepted and $4.7$\,s per declined child, lowers the speedup only from $1.34\times$ to $1.33\times$.
Time counts the model calls that completed; calls interrupted by the evaluation infrastructure and retried are excluded on every benchmark, since they reflect the environment rather than the method, although their tokens remain in the cost.
On GAIA and \datasetname{}, the FM calls of the variant check were not timed either, which slightly favors \methodname{} in their speedups.
Time also excludes script execution and the variant check outside model calls; measured by wall clock from a task's first model call to its score, which includes both as well as retries, \methodname{} is slower than the neural agent on augmented GAIA ($0.64\times$) and GAIA ($0.46\times$).
Cost efficiency and speedup are the neural agent's cost and time divided by those of \methodname{}, computed over the tasks on which both systems produced a result without an error, whether or not that result was correct; this excludes three of the $58$ paired \datasetname{} runs, which have no scored verdict, and nothing on the other benchmarks.
On GSM-Symbolic, the neural agent's per-task results were not preserved, so both ratios are computed over all tasks.
Figure~\ref{fig:free_rate} uses the same basis; Table~\ref{tab:mechanism} compares settings over all tasks except on \datasetname{} (Appendix~\ref{appdx:ablations}).
On augmented GAIA, costs are paired on the 378 child tasks, and the $165$ parent tasks cost \methodname{} a further \$$257.90$ and $70.0$ hours, so no paired total over whole families exists.
On $\tau^2$-bench, cost and time count the agent alone and exclude the benchmark's simulated user, which runs on a different model provider and is driven identically by both systems; the one-off authoring cost is likewise agent-only, so that \methodname{} pays back its authoring cost after $1.44$ evaluation passes on telecom and $2.36$ on airline.
On \datasetname{}, time is the agent's compute time over its model calls, plus the executor's replay time for \methodname{}.

\paragraph{Statistical tests.}
We use the exact McNemar test for paired binary outcomes on GAIA, augmented GAIA and the per-task $\mathrm{Pass}^4$ outcomes of $\tau^2$-bench, the Wilcoxon signed-rank test over tasks for the paired $\mathrm{Pass}^1$ difference on $\tau^2$-bench, the sign test over tasks on WorkArena L1, and the sign test over the $42$ configurations of \datasetname{}.
Throughout the paper, we treat a difference as statistically significant at $p<0.01$.
On GSM-Symbolic, per-task outcomes of the neural agent were not preserved, so we use an unpaired two-proportion test, which is conservative.
Small benchmarks and levels have little power, so the per-level results are descriptive.

\paragraph{Component ablations.}
The GSM-Symbolic ablation uses the $1000$ tasks of the $20$ templates with the lowest identifiers; its free share and cost are measured on this subset, where $96.0\%$ of tasks are free against $92.6\%$ on the full set, which is why its cost efficiency exceeds that in Table~\ref{tab:main_results}.
The augmented GAIA setting without reuse ran at an earlier code revision than the other augmented GAIA settings.
In the \datasetname{} setting with family reuse, cheap rebinding was attempted on every task but never took effect: the loop programs take their per-record values from the task's data rather than from declared parameters, and the rebinding step neither maps onto that data nor supplies it, so this setting behaves as family reuse with the variant check alone.
The \datasetname{} settings with the executor loop disabled are valid only inside this decomposition, and the setting with the loop but without the variant check lost two tasks to infrastructure faults, so its denominator is $150$ rather than $180$ records.

\paragraph{Program transfer.}
On GSM-Symbolic, each family's program is authored in instance order, so a program saved at instance $k$ has seen instances $0$ to $k$; applying it to a later instance measures forward generalization and to an earlier one backward compatibility, over $38{,}503$ substitute-and-execute trials without any model call.
In $3.85\%$ of pairs the stored program's parameter interface does not fit the target at all; we report these separately, so that coverage is $97.82\%$ over the pairs where the program applies and $96.7\%$ if every such pair is counted as a failure, and $310$ of the $315$ execution errors are interface mismatches.
Of the $100$ families, $91$ reach complete forward coverage; of the remaining nine, four fail only on a single instance that no program of the family ever solved, four show diffuse variation in program quality, and one keeps changing its parameter interface.
On augmented GAIA, the control is the same configuration with reuse disabled, and because the backward-compatibility requirement and the family regression cases are only added when family context exists, the control isolates this mechanism.
For the GSM-Symbolic comparison in Table~\ref{tab:transfer}, the reuse side is the full configuration of Table~\ref{tab:main_results} on all $100$ families, taking at each instance $k$ the version of the family's program in use at $k$, and the control is the configuration without reuse, whose programs are authored for a single instance at five randomly chosen $k$ per family.
Each control program and the corresponding stored program are applied to the same $30$ randomly chosen other instances of the family: one FM call binds the values of the target instance to the program's own parameters, and the program is then executed without further model calls and scored against the target's answer, giving $15{,}000$ pairs.
We test significance with the Wilcoxon signed-rank test over the $100$ families, since pairs within a family are not independent; the family-level difference is $+12.5$ points ($95\%$ bootstrap CI $[8.0, 17.4]$).
The two sides were run at different code revisions, and the full configuration passes a program on to later instances only if it solved its own task; restricted to targets the stored program has not seen, the difference is still $+11.6$ points.
Binding parameters by name instead, without an FM call, fails on $90\%$ of the control's pairs, since independently authored programs name their parameters differently; where the names do fit, its forward coverage is $99.8\%$ ($542$ of $543$ pairs), and Table~\ref{tab:robustness} reports both procedures.

\paragraph{Replay profile.}
We also profile the per-record replay time of \methodname{} alone, since the neural agent has no executor to profile; the pooled median slope is $+0.00029$ per record, and at $N\!=\!100$ the per-record time decreases from $1.12\times$ the task mean in the first fifth of records to $0.93\times$ in the last.

\paragraph{Adaptation on $\tau^2$-telecom.}
In Table~\ref{tab:tau2_ladder}, the second row gives the neural agent the same task transcripts that \methodname{} authors from, and the third row executes an authored policy for $96.2\%$ of its actions.
Table~\ref{tab:tau2_ladder} comes from an earlier run of the same protocol, on the same $114$ tasks with four trials each, than the runs behind Table~\ref{tab:main_results} and Table~\ref{tab:tau2}, which is why its first and third rows differ slightly from the $\mathrm{Pass}^1$ values there ($0.603$ against $0.608$, and $0.789$ against $0.816$); its first two rows also ran on a different agent harness than its last two.

\paragraph{Case study.}
The programs in Figure~\ref{fig:code_case_study} are quoted verbatim; in (a), instances \texttt{i01} and \texttt{i39} of the same GSM-Symbolic family differ only in the highlighted constants.

\section{Augmented GAIA Dataset} \label{appdx:gaia_augmented}
\subsection{Construction} \label{appdx:gaia_construction}
The validation set of the GAIA dataset \citep{mialon2023gaiabenchmarkgeneralai} consists of 165 unique computer agent tasks. It is separated into 3 levels of difficulty, with 53, 86, and 26 tasks in level 1, 2, and 3. To assess the efficacy of our neuro-symbolic approach over this type of tasks, we augment this validation set with 129, 204, and 45 new data points for level 1, 2, and 3. We call this set of 378 data points the \textbf{Augmented GAIA dataset}.
Constructing parametric variants of a fixed benchmark to separate reasoning from memorization follows GSM-Symbolic~\citep{mirzadeh2024gsmsymbolic} and Putnam-AXIOM~\citep{gulati2025putnamaxiom}.

The Augmented GAIA dataset is created through the followup pipeline.
\begin{compactitem}
    \item \textbf{Task proposer} For each data point in the original GAIA validation set, we ask an LLM to propose 3 alternative tasks that are very similar to the original task in format and potential solution, but nonetheless different in a small way. For example, for an original task of "What is the oldest Blu-Ray recorded in this spreadsheet?", the proposer is encouraged to propose tasks such as "What is the newest Blu-Ray recorded in this spreadsheet?", "What is the oldest DVD recorded in this spreadsheet?", or "What is the second oldest Blu-Ray recorded in this spreadsheet?" The proposer has access to the original question, answer, and a copy or textual representation of the file needed for the task, depending on the file types. We call the original data point the \textbf{parent} data point and all the proposed alternatives its \textbf{children} data points.

    \item \textbf{Ground Truth Generator} We use Co-Sight \citep{zhang2025cosight}, an LLM agent design that achieved SOTA performance over the test dataset of GAIA at the time of experiment\footnote{This is no longer the case. For the GAIA test set leaderboard, see https://huggingface.co/spaces/gaia-benchmark/leaderboard}, to attempt to answer the children tasks generated, with necessary files inherited from their parents. We run this answer generation process for each child task 3 times and conduct majority voting the collect the final answer that we label as ground truth. Due to the expected failures of the Co-Sight agent and infeasibility of some of the generated tasks, we do not consider this ground truth to be reliable. However, we consider them to be a good approximation of the true answer if the task is feasible, and they serve our purpose of evaluating the reusability of generated codes by the Neuro-Symbolic agent.

    \item \textbf{Filterer} As a final step, we use a simple, heuristic script to remove any data points that Co-Sight failed to find a meaningful answer to. The final result is the Augmented GAIA dataset.
\end{compactitem}

\subsection{Additional Results} \label{appdx:gaia_results}
Table~\ref{tab:gaia_results} breaks down success rates on GAIA and augmented GAIA by difficulty level.

\begin{table}[ht]
\centering
\small
\setlength{\tabcolsep}{4.5pt}
\begin{tabular}{@{}lcccc@{\hskip 12pt}cccc@{}}
\toprule
 & \multicolumn{4}{c}{GAIA} & \multicolumn{4}{c}{Augmented GAIA} \\
\cmidrule(lr){2-5} \cmidrule(lr){6-9}
System & L1 (53) & L2 (86) & L3 (26) & all (165) & L1 (129) & L2 (204) & L3 (45) & all (378) \\
\midrule
Reference (HAL) & 82.1 & 72.7 & \textit{65.4} & 74.6 & --- & --- & --- & --- \\
Neural agent & \textit{86.8} & \textit{75.6} & 57.7 & \textit{76.4} & \textit{62.0} & \textit{58.3} & \textit{57.8} & \textit{59.5} \\
\methodname{} & \textit{86.8} & 72.1 & 46.2 & 72.7 & 58.9 & 57.4 & 51.1 & 57.1 \\
\midrule
McNemar $p$-value & 1.000 & 0.648 & 0.453 & 0.377 & 0.481 & 0.851 & 0.508 & 0.281 \\
\bottomrule
\end{tabular}
\caption{Task success rate (\%) on GAIA and augmented GAIA by difficulty level, with the number of tasks in parentheses.
The reference row is the strongest generalist agent on the HAL leaderboard \citep{hal2025leaderboard}, $p$ is the exact McNemar test, and italics mark the strongest system in each column; no level differs significantly.}
\label{tab:gaia_results}
\label{tab:augmented_gaia_results}
\end{table}

\FloatBarrier %
\section{Augmented WorkArena Dataset} \label{appdx:workarena_augmented}
\subsection{Construction} \label{appdx:wacf_construction}

To better evaluate the Neuro-Symbolic agent, we create a new dataset focusing on complex enterprise tasks. In realistic enterprise tasks, human often need to perform multiple similar tasks, decide to take different task branches based on certain conditions and/or compare field values to find inconsistency. One example task is \textit{Retrieve a list of products and for each of them, check the sale amount, and return the products if their sale amount are less than 100 in this month}. As no existing such benchmark exists, we extend the WorkArena benchmark with 18 new tasks with loop, conditional branch and/or comparison. Each new task is built using the original WorkArena L1 tasks with the new task structure. Details are shown in Table \ref{table:complex-task}. Similar to the original WorkArena task, each complex task can be instantiated from a template with different parameters. For example, \textit{CreateMultipleUserTask} can be instantiated to create 2-10 different users. The average length of the golden traces is 42 steps, significantly longer than the original L1 tasks. Loop tasks are the longest, with 81 steps on average, while conditional / comparison tasks are shorter, ranging between 14-25 steps.

\begin{table}[ht]
\small
\centering
\begin{tabular}{ll}
\toprule
\textbf{Task Type} & \textbf{Task Names} \\
\midrule
Loop tasks &
CreateMultipleChangeRequestTask \\
& CreateMultipleHardwareAssetTask \\
& CreateMultipleIncidentTask \\
& CreateMultipleProblemTask \\
& CreateMultipleUserTask \\
& DeleteMultipleUserTask \\
& OffBoardMultipleUserTask \\
& OrderMultipleDevicesTask \\
\midrule
Conditional tasks &
\scriptsize EditChangeRequestScheduleConditionalTask \\
& EditHardwareConditionalTask \\
& EditIncidentConditionalTask \\
& EditProblemConditionalTask \\
& OnBoardUserConditionalTask \\
\midrule
Comparison tasks &
CompareChangeRequestPriority \\
& CompareExpenseLineAmountTask \\
& CompareHardwareDate \\
& CompareIncidentBusinessDuration \\
& CompareProblemPriority \\
\bottomrule
\end{tabular}
\caption{Complex Task types and corresponding task names}
\label{table:complex-task}
\end{table}

\subsection{Additional Results} \label{appdx:wacf_results}
\begin{figure*}[ht]
\centering
\input{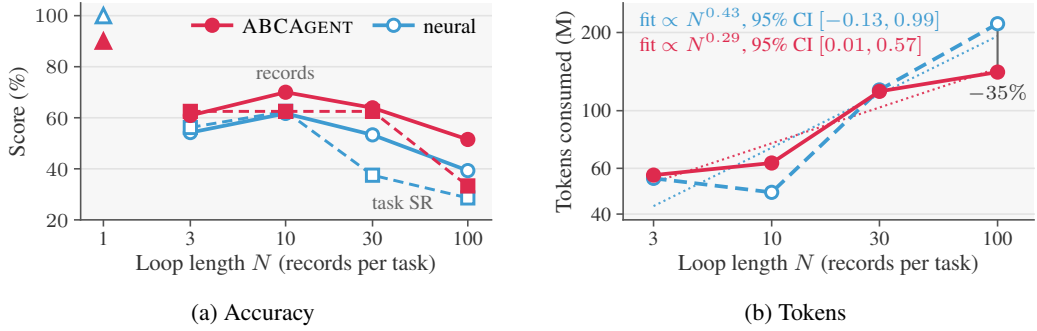}
\caption{Accuracy and token consumption on \datasetname{} as a function of loop length $N$, with the legend in (a) applying to both panels.
In (a), circles give the fraction of records written correctly and squares the task success rate, with the non-loop tasks at $N\!=\!1$; in (b), the dotted lines are least-squares power-law fits, both sublinear in $N$, labeled with $95\%$ confidence intervals on their exponents.}
\label{fig:wacf_sweep}
\label{fig:token_scaling}
\end{figure*}

Figure~\ref{fig:wacf_sweep} shows accuracy and token consumption on \datasetname{} at each loop length.
Token consumption grows sublinearly with the number of records for both systems, roughly as $N^{0.29}$ for \methodname{} ($95\%$ CI $[0.01, 0.57]$) and $N^{0.43}$ for the neural agent ($[-0.13, 0.99]$), so that \methodname{} uses $35\%$ fewer tokens at $N\!=\!100$; with four loop lengths, the two exponents are not statistically distinguishable.

\FloatBarrier
\section{Additional Results on $\tau^2$-bench} \label{appdx:tau2}
Table~\ref{tab:tau2} reports both $\mathrm{Pass}^1$ and $\mathrm{Pass}^4$ on the two $\tau^2$-bench domains, together with the paired difference between \methodname{} and the stock agent.

\begin{table}[ht]
\centering
\small
\begin{tabular}{@{}lcccc@{}}
\toprule
 & \multicolumn{2}{c}{telecom ($n\!=\!114$)} & \multicolumn{2}{c}{airline ($n\!=\!50$)} \\
\cmidrule(lr){2-3} \cmidrule(lr){4-5}
System & $\mathrm{Pass}^1$ & $\mathrm{Pass}^4$ & $\mathrm{Pass}^1$ & $\mathrm{Pass}^4$ \\
\midrule
Reference \citep{barres2025tau2bench} & 0.59 & 0.29 & 0.59 & 0.42 \\
\midrule
Stock agent & 0.6075 & 0.4737 & \textit{0.6850} & 0.5800 \\
\methodname{} & \textbf{0.8158} & \textbf{0.7193} & \textit{0.6850} & \textit{0.6000} \\
\midrule
paired $\Delta$ & \multicolumn{2}{c}{$\mathbf{+0.208}$ (46/16/52)} & \multicolumn{2}{c}{$+0.000$ (8/8/34)} \\
Wilcoxon $p$ & \multicolumn{2}{c}{\textbf{0.0001}} & \multicolumn{2}{c}{0.958} \\
\bottomrule
\end{tabular}
\caption{Performance on the two $\tau^2$-bench domains, with the benchmark's own stock agent as the model-matched neural baseline.
The paired row gives the mean per-task difference, \methodname{} minus stock, with win, loss and tie counts; the reference row is the strongest model evaluated by \citet{barres2025tau2bench} under the same stock agent.}
\label{tab:tau2}
\end{table}

\FloatBarrier

\section{Transfer Within a Family} \label{appdx:transfer}
\begin{figure}[h]
\centering
\input{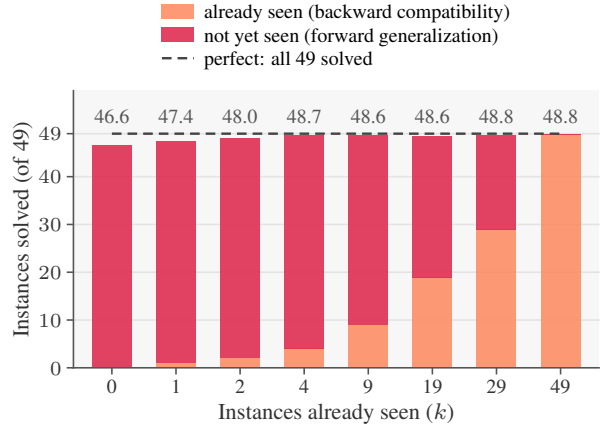}
\caption{Instances of a GSM-Symbolic family that a program frozen after seeing $k$ of its $50$ instances solves among the other $49$, split into instances it has already seen and instances it has not.
Each bar's total is shown above it, and the dashed line marks all $49$.}
\label{fig:coverage_exposure}
\end{figure}

Table~\ref{tab:robustness} extends Table~\ref{tab:transfer} with further transfer settings, together with the transfer of a frozen augmented GAIA parent to its children.
Figure~\ref{fig:coverage_exposure} shows, on all $100$ GSM-Symbolic families, how coverage of a family's other instances changes as the program sees more of them.

\begin{table}[h]
\centering
\small
\setlength{\tabcolsep}{5pt}
\begin{tabular}{@{}lrccc@{}}
\toprule
Transfer setting & pairs & Reuse & Control & $p$ \\
\midrule
\multicolumn{5}{@{}l}{\emph{Augmented GAIA}} \\
\quad All pairs & 572 & \textbf{0.388} & 0.292 & \textbf{2.2e-5} \\
\quad Forward ($j>i$) & 235 & 0.332 & 0.302 & 0.410 \\
\quad Backward ($j<i$) & 239 & \textbf{0.469} & 0.305 & \textbf{3.5e-5} \\
\quad Backward, target known-passing & 145 & \textbf{0.703} & 0.462 & \textbf{5.1e-5} \\
\quad Backward, rewritten \& known-passing & 103 & \textbf{0.718} & 0.388 & \textbf{1.2e-6} \\
\midrule
\multicolumn{5}{@{}l}{\emph{GSM-Symbolic}, parameters bound by name} \\
\quad Forward ($j>k$) & 5\,572 & \textbf{0.966} & 0.097 & $\mathbf{<10^{-4}}$ \\
\quad Backward ($j<k$) & 2\,252 & \textbf{0.972} & 0.090 & $\mathbf{<10^{-4}}$ \\
\midrule
\multicolumn{5}{@{}l}{\emph{Frozen-parent transfer}, augmented GAIA, reuse only} \\
\quad Unseen children, FM-assisted rebind & 303 & 0.271 & --- & --- \\
\quad Unseen children, own parameters & 303 & 0.241 & --- & --- \\
\quad Seen parents, own parameters & \phantom{0}75 & 0.733 & --- & --- \\
\midrule
\multicolumn{5}{@{}l}{\emph{FM-assisted rebinding}, each setting's own programs} \\
\quad GSM-Symbolic, forward ($j>k$) & 7\,640 & \textbf{0.958} & 0.842 & $\mathbf{<10^{-4}}$ \\
\quad GSM-Symbolic, backward ($j<k$) & 7\,360 & \textbf{0.963} & 0.826 & $\mathbf{<10^{-4}}$ \\
\quad Augmented GAIA, siblings & 234 & 0.303 & 0.235 & 0.020 \\
\bottomrule
\end{tabular}
\caption{Transfer of a stored program to other tasks of the same family, to tasks it has not seen (forward) and to tasks it was authored on earlier (backward).
Bold marks differences from the no-reuse control that are significant at $p<0.01$, by McNemar's test on augmented GAIA and by a sign test over the $20$ families on GSM-Symbolic, where a program whose parameters cannot be bound to the target counts as a failure.
The frozen-parent rows bind a child task either with one FM call (FM-assisted) or with the parameter values from the child's own program, and the last block binds every pair with one FM call, on GSM-Symbolic over $100$ families with the full configuration's programs, tested with the Wilcoxon test over families (Appendix~\ref{appdx:protocols}), and on augmented GAIA from each family's first child solved in both arms to its siblings.}
\label{tab:robustness}
\end{table}

\FloatBarrier

\section{Additional Analysis} \label{appdx:analysis}

\paragraph{How often programs defer to the FM, and how deeply.}
We find that programs defer a step to an FM sub-call only where the step resists code: $7.1\%$ of stored GAIA programs defer such a step ($251$ of $3{,}541$, steady at $6.3$--$10.3\%$ across $16$ runs), whereas none of the $4{,}999$ GSM-Symbolic programs do.
When a GAIA program does defer a step, it almost always uses the generic \texttt{delegate} call rather than a typed perception helper ($304$ of $313$ calls), so the deferred step is an unconstrained FM sub-call.
Although the design allows a delegated call to author a program of its own, nesting beyond one level is rare in practice: it happened once in our runs, and a third level never occurred.
Among GAIA tasks that defer a step, runs whose \texttt{delegate} call returned a result pass $75.4\%$ of the time, against $58.5\%$ where it failed ($p = 0.022$).
We do note, however, that failed calls are plausibly harder sub-questions, and that the within-task comparison has only two discordant pairs, so the effect of a successful call lies between $0$ and $+17$ points.
These calls also failed often in our runs ($2{,}231$ failures, over half of them timeouts), and since a failed call returns an empty result, this biases the reported GAIA numbers against \methodname{} if anything.

\paragraph{Failures on WorkArena L1.}
Of the $35$ failures of \methodname{} on the paired WorkArena L1 tasks, classified by the benchmark's own validator messages, $22$ end with the browser on a page other than the one the validator reads, including $19$ of the $28$ that the neural agent solved: $14$ catalog orders never reach the checkout confirmation that carries the request identifier, which appears in none of them but in all $13$ successful orders of \methodname{}, and $8$ chart and sorting tasks leave the target page.
The remaining failures are a wrong interface state ($6$), wrong record fields ($4$, three of which the neural agent also gets wrong) and wrong answers ($3$); none are execution errors.

\paragraph{Errors over loop length.}
Errors on \datasetname{} do not accumulate with loop length: $87.5$--$100\%$ of field errors fall in a field that is wrong on at least $80\%$ of the records requesting it.
Hence, what grows with $N$ is not the chance of a new mistake on each record, but the number of records that a single authoring defect is replayed on.

\paragraph{Inspectability.}
Because \methodname{} stores the program it executes, the reason a program does or does not transfer is legible in the program itself (Figure~\ref{fig:code_case_study}).
In a GSM-Symbolic family, all instances share one derivation, so a sibling is solved by rebinding a handful of constants; in an augmented GAIA family built around Boggle boards, a new board calls for a new search rather than a new constant, which is why rebinding fails there.
The neural agent, in contrast, leaves no such artifact to inspect.

\begin{figure*}[h]
\centering
\input{figures/polished/code-case-study}
\caption{Programs authored by \methodname{}, taken verbatim from run artifacts.
In (a), all instances of a GSM-Symbolic family share one derivation, so a sibling is solved by rebinding the highlighted constants; in (b), rebinding fails on an augmented GAIA family, because a new Boggle board requires a new search rather than a new constant.}
\label{fig:code_case_study}
\end{figure*}

\FloatBarrier

\section{Component Ablations by Benchmark} \label{appdx:ablations}
Table~\ref{tab:mechanism} restricts augmented GAIA to its $129$ L1 child tasks, and uses the $8$ loop composites of \datasetname{} at $N\!=\!30$ together with its $10$ non-loop tasks.
On \datasetname{}, the success rate is computed over the tasks with a score, and cost efficiency over those scored for both systems; the non-loop tasks of the setting without reuse and rebinding coincide with the full configuration by construction, since a single task has no family to reuse, and its loop tasks run without the executor loop (Table~\ref{tab:ablation_wacf}).
The setting without the variant check is cheaper than the neural agent only because its loop tasks commit records with wrong field values, none of them correct, and it also solves fewer tasks, so its cost efficiency should be read together with its success rate and records; all four settings solve every non-loop task.
Cost efficiency is colored as in Table~\ref{tab:main_results}.
On GSM-Symbolic, every setting in Table~\ref{tab:mechanism} and Table~\ref{tab:ablation_gsm} is run on the same $1{,}000$ tasks from $20$ templates and costed from token usage in a single pass; the full configuration is the corresponding slice of the $5{,}000$-task run in Table~\ref{tab:main_results}, run at a different code revision, and its binding and verification calls, which were not logged per task, are estimated at \$$3.62$ from the per-call cost of that run.
$\tau^2$-bench is omitted, since it has no task families to rebind across and its authored policy has no parameters for the variant check to vary; a task that does require authoring costs about the same for both systems, \$$1.610$ against \$$1.592$ on augmented GAIA.
Table~\ref{tab:mechanism} keeps only the component settings that were measured on more than one benchmark.
This section reports the settings that were measured on a single benchmark, which isolate one component at a time on GSM-Symbolic (Table~\ref{tab:ablation_gsm}) and the interaction between the variant check and the executor loop on \datasetname{} (Table~\ref{tab:ablation_wacf}).

\begin{table}[ht]
\centering
\small
\setlength{\tabcolsep}{5pt}
\begin{tabular}{@{}lcccc@{}}
\toprule
Setting & SR & Free & \$/task & Cost eff. \\
\midrule
Neural agent & 99.7 & 0.0\% & 0.1362 & 1.00$\times$ \\
\methodname{}, no components & \textit{99.8} & 0.0\% & 0.1408 & 0.97$\times$ \\
\quad $+$ family reuse & \textit{100.0} & 0.0\% & 0.2138 & \worse{0.64$\times$} \\
\quad $+$ variant check & 99.7 & 0.0\% & 0.2155 & \worse{0.63$\times$} \\
\quad $+$ cheap rebind & 99.1 & 96.0\% & 0.0114 & \better{11.91$\times$} \\
\bottomrule
\end{tabular}
\caption{Component ablation on GSM-Symbolic over 1000 tasks from 20 templates, adding components cumulatively.
Cheap rebinding is the only component that reduces cost, and accuracy is near ceiling for every setting.}
\label{tab:ablation_gsm}
\end{table}

\begin{table}[ht]
\centering
\small
\setlength{\tabcolsep}{6pt}
\begin{tabular}{@{}lcc@{}}
\toprule
 & executor loop off & executor loop on \\
\midrule
variant check off & \textit{64.4} & 0.0$^{\ddagger}$ \\
variant check on & 63.3 & 63.9$^{\star}$ \\
\midrule
neural agent & \multicolumn{2}{c}{53.3} \\
\bottomrule
\end{tabular}
\caption{Records written correctly (\%) on \datasetname{} at $N\!=\!30$, crossing the variant check with the executor loop, without family reuse.
$^{\star}$ is the configuration reported as \methodname{}; $^{\ddagger}$without the variant check, the replayed programs commit records with wrong field values, including one program that ignores its per-record input entirely and submits the same values on every record.}
\label{tab:ablation_wacf}
\end{table}

\end{document}